\documentclass[10pt,twocolumn,letterpaper]{article}

\usepackage{cvpr}              

\usepackage[dvipsnames]{xcolor}

\usepackage{booktabs} 
\usepackage{amsfonts}
\usepackage{enumitem}
\definecolor{cvprblue}{rgb}{0.21,0.49,0.74}
\usepackage[pagebackref,breaklinks,colorlinks,citecolor=cvprblue]{hyperref}

\def\paperID{11607} 
\def\confName{CVPR}
\def\confYear{2024}

\title{Distinctive Image Captioning: Leveraging Ground Truth Captions in CLIP Guided Reinforcement Learning}

\author{Antoine Chaffin\\
IRISA, IMATAG\\
Rennes, France\\
{\tt\small antoine.chaffin@irisa.fr}
\and
Ewa Kijak\\
IRISA\\
Rennes, France\\
{\tt\small ewa.kijak@irisa.fr}
\and
Vincent Claveau\\
IRISA, CNRS\\
Rennes, France\\
{\tt\small vincent.claveau@irisa.fr}
}

\usepackage[normalem]{ulem}

\begin{document}

\maketitle

\begin{abstract}
Training image captioning models using teacher forcing results in very generic samples, whereas more distinctive captions can be very useful in retrieval applications or to produce alternative texts describing images for accessibility. Reinforcement Learning (RL) allows to use cross-modal retrieval similarity score between the generated caption and the input image as reward to guide the training, leading to more distinctive captions. Recent studies show that pre-trained cross-modal retrieval models can be used to provide this reward, completely eliminating the need for reference captions. 
However, we argue in this paper that Ground Truth (GT) captions can still be useful in this RL framework.
We propose a new image captioning model training strategy that makes use of GT captions in different ways. Firstly, they can be used to train a simple MLP discriminator that serves as a regularization to prevent reward hacking and ensures the fluency of generated captions, resulting in a textual GAN setup extended for multimodal inputs. Secondly, they can serve as additional trajectories in the RL strategy, resulting in a teacher forcing loss weighted by the similarity of the GT to the image. This objective acts as an additional learning signal grounded to the distribution of the GT captions. Thirdly, they can serve as strong baselines when added to the pool of captions used to compute the proposed contrastive reward to reduce the variance of gradient estimate. Experiments on MS-COCO demonstrate the interest of the proposed training strategy to produce highly distinctive captions while maintaining high writing quality. 
\end{abstract}

\section{Introduction}
\label{sec:intro}

Image captioning is the task of generating a description of the semantics of an image in natural language. One major challenge in this domain is to generate \textbf{distinctive} captions, that is, a description that allows to distinguish between the input image and other (similar) ones. For instance, "one person is standing" can be considered as a correct caption for several images showing someone: it is a correct sentence that fundamentally describes such images, yet it is not describing specifically a given image more than another. In contrast to generic ones, distinctive captions are more informative and descriptive. This is an expected property for retrieval applications, by indexing images using an appropriate textual representation, or to provide further details to people with vision impairment.

Captions in standard datasets~\cite{DBLP:conf/eccv/LinMBHPRDZ14, DBLP:conf/acl/SoricutDSG18, DBLP:journals/ijcv/KrishnaZGJHKCKL17, DBLP:journals/corr/abs-2210-08402} 
only describe the most salient objects in the image, that are often common to many images. Thus, captioning models trained to match Ground Truth (GT) captions tend to generate overly generic captions, and often produce the exact same caption for different images that share the same global semantics~\cite{DBLP:conf/iccv/DaiFUL17, DBLP:conf/nips/DaiL17, DBLP:conf/cvpr/WangC19, DBLP:conf/ijcai/WangWLXLZZ16, DBLP:conf/naacl/00010KDBB22, DBLP:conf/wacv/HondaWM23}. The reason is that an easy way to optimize usual image captioning metrics based on word matching is to generate words that are common across training samples, and not to generate very specific words that are present in very few captions.

\begin{figure*}
    \begin{center}
    \includegraphics[width=\textwidth]{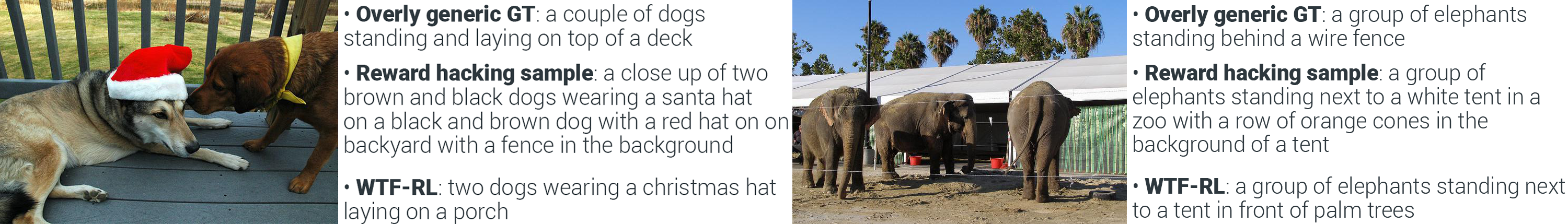}
        \caption{Examples of images with an overly generic ground truth caption, a caption generated by a model without regularization (leading to reward hacking), and the caption generated by our approach (well-written and distinctive).}
        \label{fig:captioning_cases}
    \end{center}
\end{figure*}

The distinctiveness of a caption can be measured by a cross-modal retriever: the generated text should allow to retrieve the target image among all others~\cite{DBLP:conf/nips/DaiL17, DBLP:conf/cvpr/LuoPCS18}.
A Language Model (LM) can thus be trained to generate texts that optimize the retrieval score using Reinforcement Learning (RL) by learning from generated sequences that yield high score.
Recently, advances in cross-modal retrieval models enabled the use of fixed pre-trained models such as CLIP~\cite{DBLP:conf/icml/RadfordKHRGASAM21, DBLP:journals/corr/abs-2210-08402} to guide the generator towards distinctive captions~\cite{DBLP:conf/naacl/00010KDBB22, DBLP:journals/corr/abs-2205-12630, DBLP:conf/eccv/ZhangWWX22}. 
Using a fixed cross-modal retriever reduces the risk of the generator and the retriever cooperatively converging towards something closer to a hash function rather than to natural language. However, since the retriever has not been trained to evaluate the quality of the input text, but only its relevance to the image, it may assign a very high similarity score to ill-formed sequences, and the LM will ultimately produce non-readable captions. A regularization of the generated sequences is thus still needed to avoid drifting too much from the natural language.

In this work, we propose a training method taking advantage of GT captions to optimize the trade-off between the distinctiveness and the writing quality of generated captions, illustrated in Figure~\ref{fig:captioning_cases}.
The use of cross-modal retrieval models in RL frees from the need for target reference captions, because the score of the produced sequence is computed by its similarity with the image rather than comparing it to a reference sequence. However, we argue that they can still be useful in this setup.
First, they can be used to train a simple MLP to distinguish between real and generated samples. This discriminator can replace manually defined regularization criteria from other approaches that leverage pre-trained CLIP models. This results in a GAN~\cite{DBLP:conf/nips/GoodfellowPMXWOCB14} environment, where the discriminator and the generator improve together. Second, they can be treated as generated sequences in the RL paradigm, resulting in a teacher forcing objective weighted by the similarity score of the caption to the image, thus promoting the most descriptive captions among these. This allows to learn to generate more distinctive captions using only GT captions. 
Coupling this objective with the more traditional RL one computed on samples generated by the LM allows to perform exploration while having a learning signal grounded to the human distribution that shares the same objective.
Finally, GT can further be used as candidate baselines in our proposed contrastive reward that uses the strongest baseline in a batch to reduce the variance of the gradient estimation.



Background on training distinctive image captioning models is first introduced in Section~\ref{sec:training_lm}.
Then, we present our proposed approach by introducing 1) the use of the discriminator, 2) the use of ground truth as additional trajectories and 3) the contrastive rewards in Section~\ref{sec:method}. 
Finally, we compare the results of the approach to a model trained following~\cite{DBLP:conf/naacl/00010KDBB22} and provide some insights on our different contributions through an ablation study in Section~\ref{sec:experiments}.
\section{Related Work}
\label{sec:training_lm}
\paragraph{Teacher Forcing.}

\sloppy Language modeling estimates the probability distribution of sequences of symbols $x_1, x_2, \cdots, x_T$ (called \textit{tokens}) taken from a vocabulary $\mathcal{V}$, with variable lengths $T$. 
Given a training set of texts, a LM can be trained by optimizing weights $\theta$ of a neural network which outputs a probability distribution over the dictionary for the next token given the input ones, i.e $p_\theta(x_t \mid x_{1:t-1})$ at a given time step $t$. 
As only the exact ground truth sequence is guaranteed to be right and the correctness of even very small variations of it is unknown, the model is trained to predict the next token from the GT $x^{gt}$ given previous GT tokens, by optimizing its probability through a cross-entropy loss between the target token and the output distribution. This results in the Teacher Forcing (TF) loss~\cite{DBLP:journals/neco/WilliamsZ89}: $ L(\theta)= -\sum_{t=1}^{T}  \log p_\theta(x^{gt}_{t} \mid x^{gt}_{1:t-1})$. 
A basic approach to train an image captioning model is to train a LM to produce the caption $x^{gt}$ while being conditioned to the input image $i$ using teacher forcing. The image can be seen as additional previous tokens used as context, resulting in a very similar loss: $L_\theta(x^{gt})= -\sum_{t=1}^{T} \log p_\theta(x^{gt}_{t} \mid x^{gt}_{1:t-1}, i)$.

Image captioning, when trained through TF, suffers from the same issues as any text generation task. Firstly, the exposure bias~\cite{DBLP:journals/corr/RanzatoCAZ15} induced by the mismatch between the training and the generation process. The model is never exposed to its mistakes during training but will suffer from error accumulation at test time. Secondly, TF only considers one target sequence, whereas many different sequences can represent the same semantic content and be valid targets. Finally, the loss is defined at the token level, while the quality of a sample is defined at the finished sequence level.

\paragraph{Reinforcement Learning.}

\sloppy One way to overcome the limitations of TF is to directly optimize a sequence level evaluation metric through RL~\cite{DBLP:conf/aaai/YuZWY17, DBLP:journals/corr/RanzatoCAZ15}. This objective can be any standard NLP metric such as BLEU~\cite{DBLP:journals/corr/WuSCLNMKCGMKSJL16} or ROUGE~\cite{DBLP:conf/iclr/PaulusXS18}. 
In the context of image captioning, the metric commonly optimized~\cite{DBLP:conf/cvpr/RennieMMRG17, DBLP:conf/icml/WangYMLBLMZZY22, DBLP:conf/eccv/Li0LZHZWH0WCG20, DBLP:conf/cvpr/ZhangLHY0WCG21, Hu_2022_CVPR} is CIDEr~\cite{DBLP:conf/cvpr/VedantamZP15}. These metrics are computed at the sequence level by comparing sampled sequences to GT references. Thus, they are non-differentiable and are optimized through the REINFORCE algorithm~\cite{DBLP:journals/ml/Williams92}. REINFORCE estimates the gradient by sampling sequences from the model. The LM is then trained to optimize the log likelihood of the best ones by scaling the associated gradient based on the obtained reward (e.g., the CIDEr score). 
For a generator parameterized by $\theta$, a generated sequence (Monte-Carlo sample) $x$ and its reward $r(x)$, the gradient becomes: $\nabla_\theta L_\theta(x)=-r\left(x\right) \nabla_\theta \log p_\theta\left(x\right)$.

\sloppy A \textit{baseline} $b$ can be subtracted from the reward to reduce the variance of the gradient estimate, as long as it does not depend on the sample $x$ (so the expected gradient is the same): $\nabla_\theta L_\theta(x)=-(r\left(x\right)-b) \nabla_\theta \log p_\theta\left(x\right)$. Self-Critical Sequence Training (SCST)~\cite{DBLP:conf/cvpr/RennieMMRG17} is the most widely used method for training image captioning model. SCST is an extension of REINFORCE that uses the model itself as a baseline to normalize the rewards. During the training, the current model will be used to generate a sequence $\hat{x}$ using test-time decoding method (e.g. Greedy Search (GS)) and uses its reward as a baseline ($b = r(\hat{x})$) for a sequence generated using a better decoding method, such as Beam Search (BS). 
The model probabilities of samples that are better than the actual model will be increased and the ones of samples that are worse will be decreased.
Hence, SCST optimizes a sequence evaluation metric as REINFORCE, but strongly reduces the variance induced by sampling a full sequence while also avoiding to learn a critic~\cite{DBLP:books/lib/SuttonB98, DBLP:conf/icml/MnihBMGLHSK16} that estimates the expected future reward for a given sub-sequence.

However, BLEU, ROUGE or CIDEr metrics are not totally correlated with human judgment and optimizing them directly might lead to biased results rather than human-like ones~\cite{DBLP:conf/emnlp/NovikovaDCR17}. A less biased metric is the score of a discriminator trained to differentiate the distributions of generated versus real texts. Guiding the generator towards a distribution that is indistinguishable from the real one would result in a perfect generator. Generative Adversarial Networks (GANs)~\cite{DBLP:conf/nips/GoodfellowPMXWOCB14} allowed massive improvements in generative tasks such as images generation, thanks to their capacity to approximate continuous data distribution. However, for discrete data such as text, the gradient flow cannot be back-propagated from the discriminator to the generator, therefore the problem is commonly cast as a reinforcement learning problem using scores of the discriminator as rewards~\cite{DBLP:conf/aaai/YuZWY17, DBLP:journals/corr/RanzatoCAZ15}. These scores serve as a learning signal which is not affected by the exposure bias and does not rely on manually designed metrics that can be biased to evaluate the quality of a sample.


\paragraph{Distinctive image captioning.}

Contrastive Learning for Image Captioning~\cite{DBLP:conf/nips/DaiL17} introduces the distinctiveness property of image captioning models. The generator is trained using the log-ratio of probabilities of the model with respect to a reference baseline model on positive and negative pairs created by randomly swapping the positive pairs. The goal of the model is to assign higher probabilities to positive pairs (respectively, lower to negative ones) than the reference model. 
Our approach, on the other hand, does not directly work on sequences probabilities, that are optimized using reinforcement learning as a proxy. Leveraging a dual encoder model (CLIP) allows to compute scores for every pair in the batch and to consider much more couples than what would be tractable by evaluating the conditional probability of every sequence for this model.~\cite{DBLP:conf/naacl/00010KDBB22} use CLIP score in the SCST framework to train the model and fine-tune its text encoder to detect grammatical mistakes in order to prevent reward hacking.~\cite{DBLP:conf/eccv/ZhangWWX22} improves over SCST by replacing the self-critical baseline with the CLIP similarity of the generated caption to a group of similar images and add a CIDEr reward to prevent the model from diverging. These rewards only focuse on either text-to-image or image-to-text retrieval, whereas our approach considers a whole batch of similar captions and images, thus considering both directions. Moreover, rather than a fixed grammar network/CIDEr score, we use an evolving discriminator which adapts to the generator and prevents emerging behaviors that are not observable at the sequence level (e.g., low diversity that can only be measured from a set of generated sequences). 
\cite{DBLP:conf/wacv/HondaWM23} identifies the limited vocabulary of a RL-trained model as a bottleneck for discriminativeness, preventing the model from using low-frequency words that are needed to correctly distinguish one image from another. This vocabulary collapse appears because only words sampled by the model obtain rewards, causing less frequent words to be less and less frequent~\cite{Choshen2020On}. ~\cite{DBLP:conf/cvpr/WangC19} shows that using teacher forcing during the RL training limits the vocabulary collapse. Our proposed weighted teacher forcing combines the TF and RL objectives, by 
using ground truth captions which contain words that might not be sampled by the model for the RL objective.

\section{Method}
\label{sec:method}

\begin{figure*}
    \begin{center}
    \includegraphics[width=\textwidth]{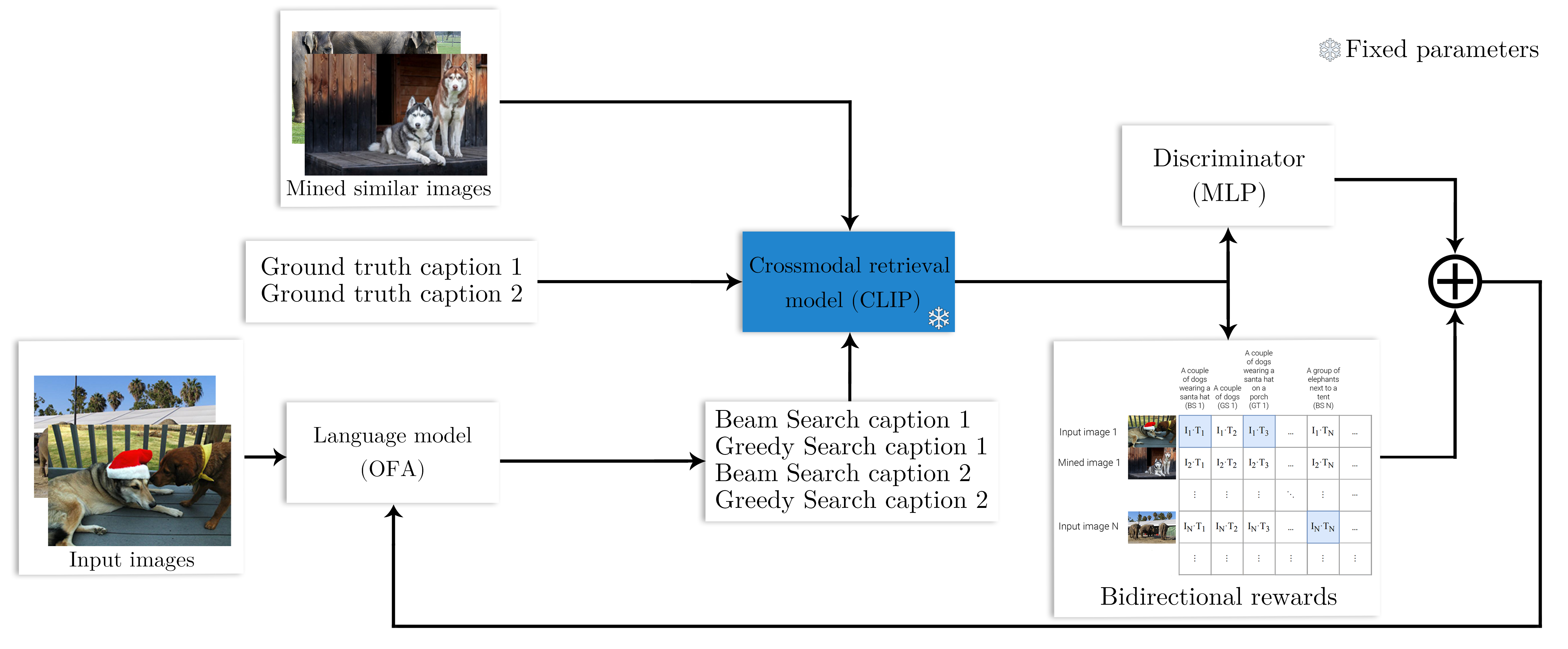}
       \caption{Proposed captioning model learning overview. Generated and ground-truth captions, as well as input and mined similar images, are projected in the CLIP embedding space. Those representations are used to compute the reward composed of a discriminator score (Section~\ref{ssec:discriminator}) and a CLIP-based bidirectional contrastive similarity score (Section~\ref{ssec:contrastive_reward}), for \textit{beam search} and \textit{ground-truth} samples (Section~\ref{ssec:wtf}) (in blue in the reward computation bloc).}
        \label{fig:approach}
    \end{center}
\end{figure*}

The proposed method extends the RL paradigm that uses the similarity score between a generated caption and its image from the cross-modal retriever CLIP as the reward. 

$f^I(i)$ and $f^T(x)$ are respectively the CLIP embeddings of image $i$ and text sequence $x$. The reward $r(x)$ is a trade-off between $r_{sim}(x)$, the similarity of $f^T(x)$ with $f^I(i)$, and $r_{regu}(x)$, a regularization based on the writing quality of the sample, controlled by a parameter $\alpha$. 
The gradient is then given by:
\vspace{0.15em}
\begin{align}
\begin{split}
\label{eqn:grad_reward}
&\nabla_\theta L_\theta(x)=-r(x) \nabla_\theta \log p_\theta(x) \\  \;\; &\mbox{ with } \; r(x)= \alpha \; r_{sim}(x) + (1-\alpha) \; r_{regu}(x)  
\end{split}
\end{align}

\subsection{Overview}
\label{ssec:overview}

The proposed learning scheme is depicted in Figure~\ref{fig:approach}. For each image $i$ in a batch, associated with its ground truth caption $x^{gt}$, similar images are mined and generated captions $x^{bs}$ and $x^{gs}$ are sampled from the LM using respectively beam search and greedy search decoding. All captions and images are projected in the CLIP embedding space. Those representations then are used to compute both terms $r_{sim}$ and $r_{regu}$ of the reward~(\ref{eqn:grad_reward}).
This reward is computed for BS samples $x^{bs}$, while GS samples $x^{gs}$ and mined images serve as baselines in the $r_{sim}$ computation.

The GT captions $x^{gt}$ are leveraged:
\begin{enumerate}[noitemsep]
    \item[(i)] to train a simple discriminator $D$ in the CLIP embedding space that discriminates between GT samples and the LM-generated BS ones, used as the regularization term $r_{regu}$ (Section~\ref{ssec:discriminator}),
    \item[(ii)] as additional training samples using the RL objective (Section~\ref{ssec:wtf}),
    \item[(iii)] as candidate baselines for the reward $r_{sim}$ (Section~\ref{ssec:contrastive_reward}).
\end{enumerate}

\subsection{Preventing Reward Hacking Using a Discriminator}
\label{ssec:discriminator}

During the exploration of the space by the policy (the LM), bad sequences with high rewards might be produced, for example by repeating important keywords for CLIP (reward hacking), or producing out-of-domain sequences 
that obtain near-random rewards. When ill-formed sequences get high scores, the model learns to reproduce them more and more, 
until the model fully collapses on this bad distribution (see Figure~\ref{fig:captioning_cases}).

To prevent learning from such ill-formed solutions, previous approaches~\cite{DBLP:conf/naacl/00010KDBB22, DBLP:journals/corr/abs-2205-12630, DBLP:conf/eccv/ZhangWWX22} use different criteria to regularize the training: detection of repetitions and grammatical errors, divergence from the distribution of the original LM, or CIDEr value. However, we argue that all of these criteria can be encapsulated in a single discriminator $D$ trained to discriminate between GT samples and the LM-generated ones. A very simple MLP classifier taking as input the representations computed by CLIP easily achieves a very high discrimination accuracy. 

The probability $P_D(x)$ of a sequence $x$ to be from a human given by~$D$ can be used as $r_{regu}(x)$. It is worth noting that, contrary to the grammar head of~\cite{DBLP:conf/naacl/00010KDBB22}, the discriminator is trained without fine-tuning the text encoder of the CLIP model, preventing a mismatch between optimizing the retrieval model used for training and the one used at test-time.
Although being less discussed in the literature, the discriminator is also useful to prevent the LM from degenerating in another situation: a well-written caption but insufficiently specific to the image (Figure~\ref{fig:captioning_cases}). Such captions will obtain negative rewards because CLIP scores them poorly, but lowering their likelihood may lead the model to unlearn the grammar and correct sequence structure. While positive rewards attract the model directly toward the trajectory, negative rewards push it away in an unknown direction. It causes the model to produce random sequences that might make the model totally collapse. The weight of the discriminator in the reward should thus be large enough to prevent both unlearning well-written texts and learning from reward hacking samples. 

Compared to the grammar network of~\cite{DBLP:conf/naacl/00010KDBB22} or the CIDEr score of~\cite{DBLP:conf/eccv/ZhangWWX22}, our discriminator not only gives information at the sentence level, but more generally on the distribution of the generated texts and can adapt to emergent behaviors of the language model that could trick a fixed model. 



\subsection{Reward-Weighted Teacher Forcing}
\label{ssec:wtf}
Since the CLIP representations of the ground truth captions are computed to train the discriminator, we propose to further leverage them to train the generative model.
Reinforcement learning scores trajectories (sequences of words in context of text generation) and learns from those that scored well. Good sampled sequences are thus required so that the model can learn from them. While this exploration process allows to find good solutions, it has a great variance and can lead to degenerate solutions (reward hacking).

GT captions can be considered a great source of relatively good solutions. We thus propose to use these captions as additional trajectories for the RL loss. 
If the reward is directly derived from the ground truth as in reference-based metrics (BLEU, ROUGE, CIDEr...), GT trajectories always obtain the upper-bound value of the reward metric. This reward is the same for every GT sequence, resulting in the standard teacher forcing objective with a learning rate multiplied by this constant. 
In our case, the cross-modal similarity score $r_{sim}$ associated to the GT is not constant (some GT captions are closer to their images in the cross-modal space). The resulting loss is thus equivalent to the teacher forcing loss weighted by the reward $r(x^{gt})$. We refer to this objective as \textbf{Weighted Teacher Forcing (WTF)}. Since the representations of ground truth captions are already computed to train the discriminator, their associated rewards are obtained through cheap dot products.


Compared to teacher forcing, the model still learns to reproduce human-written sequences but focuses more on the captions that are highly descriptive of their image, allowing to produce distinctive captions. Moreover, since these trajectories are written by humans, it strongly reduces the risk of reward hacking. Besides helping the model to stay close to the human distribution, it also enables to get back to it if the model reaches a pitfall (such as reward hacking or divergence), allowing the model to recover and start learning again. This is a very handful property since RL training is known to be unstable.

\subsection{Beyond a Single Baseline: The Bidirectional Contrastive Reward}
\label{ssec:contrastive_reward}
Previous studies use either only texts~\cite{DBLP:conf/naacl/00010KDBB22} or images~\cite{DBLP:conf/eccv/ZhangWWX22} as the baseline and thus only consider one cross-modal research direction. Inspired by the contrastive loss used to train CLIP, defined for a given couple $(x_c, i_c)$, a temperature parameter $\tau$ and a collection of negative texts $\mathcal{T}$ and negative images $\mathcal{I}$, we derive a bidirectional contrastive reward, used as similarity reward $r_{sim}$ in~(\ref{eqn:grad_reward}):
\begin{multline}
\label{eqn:bicont} 
\displaystyle
r_{bicont}(x_c)  =  r_{i2t}(x_c) + r_{t2i}(x_c) \\
  =  \tau \left( \log\frac{e^{\frac{f^T(x_c) \cdot f^I(i_c)}{\tau}}}{\sum_{x\in \mathcal{T} \setminus x_c }{e^\frac{f^T(x) \cdot f^I(i_c)}{\tau}}} \right. \\
  \left. +   \log\frac{e^{\frac{f^T(x_c) \cdot f^I(i_c)}{\tau}}}{\sum_{i \in \mathcal{I} \setminus i_c}{e^\frac{f^T(x_c) \cdot f^I(i)}{\tau}}} \right) 
\end{multline}


Please note that the reward more precisely corresponds to the definition of the decoupled contrastive loss of~\cite{DBLP:conf/eccv/YehHHLCL22} that excludes the positive couple from the denominator. $\mathcal{T}$ is composed of $x^{bs}$, $x^{gs}$, and $x^{gt}$ for all images of the batch (see Figure~\ref{fig:approach}). 
$\mathcal{I}$ is composed of all the images of the batch, as well as similar images mined in the dataset and added to the batch as additional images, following the setup of~\cite{DBLP:conf/eccv/ZhangWWX22}.
Note that although a text-to-image model as OFA~\cite{DBLP:conf/icml/WangYMLBLMZZY22} can be used to generate negative images, in practice this method is too computationally expensive and may lead to less hard negatives in CLIP space.
The mining of similar images and the computation of all image representations $f^I(i)$ is done only once before the training, and thus does not bring any computational overhead during training. As the representations $f^T(x^{gt})$ and $f^T(x^{bs})$ of GT and BS captions are already computed for the discriminator, computing the contrastive reward is thus inexpensive since it mainly consists of dot products.

Unlike previous approaches, our contrastive reward considers both directions (by normalizing either by the similarity of the image with all the captions of the batch -–image-to-text reward $r_{i2t}$–-, or by the similarity of the caption with all the images in the batch –-text-to-image reward $r_{t2i}$–-), making sure that the caption is very descriptive of the image and this image only.
Both parts of the reward can be rewritten as the similarity of the couple $(x_c, i_c)$ minus the LogSumExp (LSE) of the similarities within the batch. With a small enough temperature parameter $\tau$, LSE is an approximation of the max operator:

\begin{align*}
    &r_{i2t}(x_c)=\tau \left(\log\frac{e^{\frac{f^T(x_c) \cdot f^I(i_c)}{\tau}}}{\sum_{x \in \mathcal{T}\setminus x_c}{e^\frac{f^T(t) \cdot f^I(i_c)}{\tau}}}\right)\\
    &= \tau \left( \log(e^{\frac{f^T(x_c) \cdot f^I(i_c)}{\tau}})- \log(\sum_{x \in \mathcal{T}\setminus x_c} e^\frac{f^T(x) \cdot f^I(i_c)}{\tau}) \right) \\
    & \approx f^T(x_c) \cdot f^I(i_c) - \max_{x \in \mathcal{T}\setminus x_c}\{f^T(x) \cdot f^I(i_c)\}
\end{align*}

This image-to-text reward ($r_{i2t}$) can therefore be seen as the standard SCST where the baseline $b$ is the hardest negative: the most similar caption to the image $i_c$ among negative samples $\mathcal{T}\setminus x_c$. This motivates the use of the decoupled contrastive loss: not excluding $x_c$ from the denominator would lead to a reward that is always negative or zero, even when $x_c$ is the most similar caption to $i_c$ among every caption in the batch (the goal of the model). Please note that it reduces to the standard SCST reward of~\cite{DBLP:conf/naacl/00010KDBB22} when the most similar caption in the batch is the caption generated for the image using GS.

Applied to the text-to-image reward ($r_{t2i}$), this approximation is almost equivalent to the $G_{min}$ formulation in~\cite{DBLP:conf/eccv/ZhangWWX22} that uses the similarity of the most similar image as the baseline.
The proposed bidirectional contrastive reward thus seamlessly handles both cross-modal retrieval directions and selects the strongest baselines among a large batch, at a very low cost.
The mean similarity in the batch 
is closer to the running average, often used as a baseline in traditional RL. However, early experiments showed that it is not a strong enough baseline to prevent the model from diverging. The proposed reward results in a more conservative learning, only letting the model learn from very good sequences.



\vspace{0.2cm}
Finally, applying~(\ref{eqn:grad_reward}) with the reward defined in~(\ref{eqn:bicont}) as $r_{sim}$ and the discriminator as $r_{regu}$ to both beam search generated sequences (traditional RL) and ground truth captions (weighted teacher forcing) captions, we end up with the following gradient for a given image:
\begin{equation}
\begin{split}
\label{eqn:final_grad_reward}
&\nabla_\theta L_\theta(x^{bs}, x^{gt})= \\
&-  \biggl[ r(x^{bs}) \nabla_\theta \log p_\theta(x^{bs}) 
  + r(x^{gt}) \nabla_\theta \log p_\theta(x^{gt}) \biggr] \\
&   \;\; \mbox{ with } \; r(x)= \alpha \; r_{bicont}(x) + (1-\alpha) \; p_D(x)  
\end{split}
\end{equation}


\section{Experiments}
\label{sec:experiments}


Following previous studies~\cite{DBLP:conf/naacl/00010KDBB22,DBLP:conf/eccv/ZhangWWX22}, we measure the retrieval rate achieved using generated captions and a fixed retriever as well as their writing quality on the MS COCO dataset~\cite{DBLP:conf/eccv/LinMBHPRDZ14} using the Karpathy splits~\cite{DBLP:journals/pami/KarpathyF17}. To evaluate the contribution of each component of the proposed learning scheme, we train different variants of the proposed setup (Section~\ref{ssec:models}).

\subsection{Setup}

\paragraph{Training}
We use the state-of-the-art captioning model \href{https://huggingface.co/OFA-Sys/ofa-tiny}{OFA}~\cite{DBLP:conf/icml/WangYMLBLMZZY22} in its tiny version as LM. All the models are trained starting from the same checkpoint: tiny-OFA trained using TF for 2 epochs, for which we also report the results as baseline (\textbf{TF}). The models are then trained for 5 epochs, using a learning rate of $1e-6$, $\alpha$ set to 0.94, and a batch size of 20 ground truth caption-image pairs.

\paragraph{Discriminator}
The discriminator is a 3-layer MLP and is first pre-trained on the MS COCO train set to distinguish $x^{gt}$ and $x^{bs}$ generated by the original LM pre-trained with TF. Indeed, the discriminator should be good enough to correctly guide the LM at the beginning of the training. It is then trained throughout the LM training process, at the beginning of each iteration along with the generator, on the batch samples. 

\paragraph{Metrics}
Different metrics are used to compare different properties of generated samples.
The Recall@k metric using the fixed pre-trained CLIP model (R@k) evaluates the discriminativeness of the generated caption. This metric is reported for $k \in \{1, 5, 10\}$ and for both text-to-image and image-to-text retrieval (contrary to previous approaches that only report one cross-modal retrieval direction). Next, standard COCO captioning metrics that evaluate the writing quality are reported, including BLEU~\cite{DBLP:journals/corr/WuSCLNMKCGMKSJL16}, ROUGE~\cite{DBLP:conf/iclr/PaulusXS18}, CIDEr~\cite{DBLP:conf/cvpr/VedantamZP15}, METEOR~\cite{DBLP:conf/acl/BanerjeeL05} and SPICE~\cite{DBLP:conf/eccv/AndersonFJG16}.
The Self-BLEU~\cite{DBLP:conf/sigir/ZhuLZGZWY18} metric, corresponding to the BLEU~\cite{DBLP:conf/acl/PapineniRWZ02} metric using other generated captions as reference, is also reported. A high Self-BLEU indicates a high overlap between generated samples, implying a low diversity.

\subsection{Ablation Study}
\label{ssec:models}

We trained three variants of the proposed setup, that leverages a discriminator $D$ and uses the bidirectional contrastive reward $r_{bicont}$:

\begin{itemize}[noitemsep]
    \item \textbf{WTF-RL} uses both BS and GT trajectories, which corresponds to the policy gradient given by~(\ref{eqn:final_grad_reward}) $\nabla_\theta L_\theta(x^{bs}, x^{gt})$,
    \item \textbf{WTF} uses only GT trajectories, with policy gradient $\nabla_\theta L_\theta(x^{gt})$, 
    \item \textbf{RL} uses only generated BS trajectories, with policy gradient $\nabla_\theta L_\theta(x^{bs})$.
    
\end{itemize}

\vspace{0.2cm}
The gain brought by the bidirectional reward is studied by removing the text-to-image reward from the contrastive reward (\textbf{RL - Unidirectional}). The gradient policy is then: $$\nabla_\theta L_\theta(x^{bs})=-(\alpha \; r_{i2t}(x^{bs}) + (1-\alpha) \; p_D(x^{bs})) \log p_\theta(x^{bs}).$$
This reward is very similar to SCST with a baseline corresponding to the caption that has the highest similarity with the image in the batch, instead of considering only the GS sample as the usual SCST. These models are compared to the training setup of~\cite{DBLP:conf/naacl/00010KDBB22}, using the grammar network provided by the authors and the same weighting between the grammar and CLIP score reward (\textbf{SCST - Grammar}). To evaluate the benefits of the discriminator, we also train a model using only the GS caption as the baseline (\textbf{SCST - Discriminator}). This last setup is the same as~\cite{DBLP:conf/naacl/00010KDBB22} (SCST - Grammar), but using a discriminator rather than the grammar network.

\begin{table*}[th]
    \centering
    \resizebox{\textwidth}{!}{
    \begin{tabular}{lcccccccccccc}
          &    \multicolumn{3}{c}{\textsc{T2I retrieval}} & \multicolumn{3}{c}{\textsc{I2T retrieval}} &   \multicolumn{5}{c}{\textsc{Writing Quality}} &   \multicolumn{1}{c}{\textsc{Diversity}}   \\ 
         \cmidrule(r){2-4} \cmidrule(r){5-7}  \cmidrule(r){8-12} \cmidrule(r){13-13} 
          & R@1 ↑ & R@5 ↑ & R@10 ↑ & R@1 ↑ & R@5 ↑ & R@10 ↑ & B4 ↑ & R-L ↑ & C ↑ & M ↑ & S ↑ & Self-BLEU ↓ \\
        \midrule
        TF & 17.14 & 39.06 & 51.14 & 23.98 & 49.72 & 61.94 & 32.73 & 55.43 & 109 & 27.19 & 20.69 & 70.49 \\
        \midrule
        WTF & 20.52 & 44.58 & 57.66 & 29.32 & 56.72 & 69.08 & \textbf{32.9} & \textbf{55.57} & \textbf{110.2} & \textbf{27.46} & \textbf{21.26} &  61.45 \\
        \midrule
        WTF-RL & 33.82 & 61.98 & 73.68 & 44.26 & 73.34 & 83.4 & 24.61 & 51.05 & 86.22 & 25.7 & 20.09 & \textbf{57.55} \\
        \midrule
        RL & \textbf{35.24} & \textbf{62.9} & \textbf{75.3} & 46.68 & 75.28 & 84.66 & 21.59 & 49 & 76.06 & 25.01 & 19.21 & 58.01 \\
        \midrule
        RL - Unidirectional & 31.52 & 58.34 & 71.04 & 45.86 & 74.4 & 83.4 & 21.45 & 48.14 & 78.53 & 24.75 & 19.83 & 62.3 \\
        \midrule
        SCST - Discriminator & 34.72 & 62.46 & 74.22 & \textbf{51.38} & \textbf{79.08} & \textbf{87.54} & 16.54 & 44.62 & 46.21 & 24.31 & 18.46 & 68.88 \\
        \midrule
        SCST - Grammar & 31.84 & 58.98 & 71.10 & 44.0 & 71.86 & 81.92 & 16.35 & 45.23 & 41.24 & 25.31 & 19.72 &  80.66 \\
        \midrule
    \end{tabular}}
    \caption{Captioning results on the MS COCO dataset (Karpathy splits). R@k correspond to the retrieval rate at k using the fixed CLIP model either using text queries (T2I) or image queries (I2T). Writing quality metrics includes BLEU@4 (B4), ROUGE-L (R-L), CIDEr (C), METEOR (M) and SPICE (S). The Self-BLEU metric measures the diversity.}.
    \label{tab:wtf_rl_results}
    \vspace{-0.25cm}
\end{table*}

Although our results are not directly comparable to the ones of previous approaches due to the difference in the backbone generative model, please note that SCST - Grammar corresponds to the setup of~\cite{DBLP:conf/naacl/00010KDBB22} and that the $r_{t2i}$ reward subsume the reward of~\cite{DBLP:conf/eccv/ZhangWWX22}. This allows to contextualize the results w.r.t the current state-of-the-art.


\subsection{Results}

The results of the different models reported in Table~\ref{tab:wtf_rl_results} give information about the impact of our different contributions.

\paragraph{Use of GT as trajectories for the RL} The first observable finding is that the WTF objective alone improves retrieval metrics over TF using only GT captions, without degrading the writing quality of the model. This shows that learning from the most distinctive human-written captions allows the LM to generate captions containing more important details while staying close to the distribution of the GT captions. It is thus a better objective than TF to couple with the traditional RL one. Using GT acts as an additional regularization and prevents vocabulary collapse, while allowing to recover if the model reaches a pitfall during RL training. The combination of the two objectives (WTF-RL) results in a model that achieves competitive retrieval results while maintaining high writing quality.

\paragraph{Discriminator} Additionally, the computed representations of the GT allow to use a discriminator to replace the grammar network of~\cite{DBLP:conf/naacl/00010KDBB22}. This single replacement (SCST - Grammar/Discriminator) allows to achieve significantly higher retrieval rates without degrading the writing quality, resulting in a better trade-off. Besides, it also yields substantially lower Self-BLEU scores. Indeed, CLIP assigns great rewards to some stereotypical information structures such as "in the background" or "in the foreground". Since these sequences are grammatically correct, they are not penalized by the grammar network. The discriminator, however, adapts to the LM and learns that they are a useful clue to detect generated samples; so, it prevents the generator from generating such sequences too often. As previously mentioned, the discriminator has a global view of generated texts. It can thus detect some bad behaviors that are not visible at the sequence level and adapt to these emergent behaviors. 

\paragraph{Bidirectional reward} When using only a unidirectional image-to-text reward (RL - Unidirectional), the text-to-image retrieval metrics significantly drop compared to the proposed bidirectional reward (RL). This means that the generated captions are more descriptive of other images in the dataset than the input image. This is because, during the training, the model has not been trained to generate a caption that is more descriptive of the input image than of a similar image. This illustrates that considering both retrieval directions in the reward is needed to produce a caption that is highly descriptive of a specific image only.

\paragraph{Stronger baseline} Finally, the image-to-text reward only using the GS baseline (SCST - Discriminator) results in higher retrieval results but lower writing quality compared to using multiple textual baselines. Indeed, selecting the strongest baseline in the batch (RL - Unidirectional) lowers the similarity part of the reward ($r_{sim}$), resulting, for a fixed $\alpha$, in a higher weight of the discriminator. Although we can not conclude from these results that using a stronger baseline yields a better retrieval/writing quality trade-off, it is expected to reduce the variance even more and prevent the model from committing too early. This can help prevent the exploitation of the reward model biases, especially in the early stage of the training, where the GS samples can be very weak. We recall that this does not bring computational overhead.

\section{Conclusion}

We studied how ground truth captions can be used in a reinforcement learning training that leverages a pre-trained cross-modal retrieval model in which they are no longer required. These captions can be used as additional trajectories for the RL objective, resulting in a weighted teacher forcing objective that allows to ground the exploration to the human distribution. This additional regularization could show very useful in a setup where the retrieval model is not fixed, by forcing the model to use original captions and their vocabulary while sharing the objective of generating distinctive captions. It also allows the model to recover from the inherent instabilities of RL training.
They also can be used to train a discriminator that will ground the exploration made by the policy to the human distribution by favoring human-like generated samples. This signal serves as a regularization of the writing quality of the model, subsuming sequence-level criteria used in previous studies while preventing the emergence of bad behaviors that are not observable at the sequence level. 

Finally, we leverage the fact that dual encoder models can compute the score of every pair in the batch at a low cost, and we use the definition of the decoupled contrastive loss to select the strongest baseline in the batch. This contrastive reward, in addition to being very similar to the original training setup of the reward model, can be used in both cross-modal retrieval directions, which is important to build truly distinctive captions.
Our findings pave the way for studies that try to also improve the CLIP model jointly with the captioning model. Starting from a strong pre-trained cross-modal retriever and strongly grounding the learning to ground truth captions might help to overcome the drifting inherent of the collaboration between the two models. To enable such extensions, the code of our approach is made~\href{https://github.com/NohTow/WTF-RL}{publicly available}.


{
    \small
    \bibliographystyle{ieeenat_fullname}
    \bibliography{egbib}
}

\end{document}